\documentclass[twoside,11pt]{article}

\usepackage[abbrvbib, preprint]{jmlr2e}

\usepackage{jmlr2e}
\usepackage{amsmath}
\usepackage{multirow}
\usepackage{natbib}
\usepackage{hyperref}
\usepackage{paralist}

\pltopsep=\medskipamount
\usepackage{subcaption}

\usepackage{relsize}
\newcommand\cpp{C\nolinebreak[4]\hspace{-.05em}\raisebox{.4ex}{\relsize{-3}{\textbf{++}}} }

\usepackage[svgnames]{xcolor}
\usepackage{listings}
\newcommand{\Rdimtools}{\texttt{Rdimtools} }

\jmlrheading{volume}{2020}{pages}{date submitted}{date published}{paper id}{Kisung You}

\ShortHeadings{Rdimtools for Dimension Reduction and Estimation in R}{You}
\firstpageno{1}

\begin{document}

\title{Rdimtools: An R package for Dimension Reduction and Intrinsic Dimension Estimation}

\author{\name Kisung You \email kyou@nd.edu \\
	\addr Department of Applied and Computational Mathematics and Statistics \\ 
	University of Notre Dame \\
	Notre Dame, IN 46556, USA}


\editor{Not assigned}

\maketitle

\begin{abstract}

Discovering patterns of the complex high-dimensional data is a long-standing problem. \textit{Dimension Reduction} (DR) and \textit{Intrinsic Dimension Estimation} (IDE) are two fundamental thematic programs that facilitate geometric understanding of the data. We present \Rdimtools - an R package that supports 133 DR and 17 IDE algorithms whose extent makes multifaceted scrutiny of the data in one place easier. \Rdimtools is distributed under the MIT license and is accessible from CRAN, GitHub, and its package website, all of which deliver instruction for installation, self-contained examples, and API documentation.
\end{abstract}


\section{Introduction}

Scientists and practitioners of today often face high-dimensional data. The primary objective of data analysis would be to find patterns and gain understanding behind what is observed. When the data dimension exceeds the scope of human perception, it is mandated to rely upon some algorithms that can extract information into fathomable forms. Dimension reduction (DR) is one approach to discover structure in high-dimensional data that has long been and is still a major research program with a vast literature  \citep{engel_survey_2012a, ma_review_2013a}. DR methods explore low-dimensional structure embedded in high-dimensional space, which makes them appealing procedures for data visualization as well as a preliminary step for statistical analysis \citep{jolliffe_note_1982, mckeown_independent_2003}. Another core instrument in high-dimensional data analysis is intrinsic dimension estimation (IDE). As its name suggests, IDE tries to estimate the true dimensionality of a low-dimensional structure from which the observed data is generated \citep{camastra_intrinsic_2016a}. 

We present an R package \Rdimtools (version 0.1.2) that implements 133 DR and 17 IDE algorithms at an unprecedented scale. Each algorithm is designed to reveal certain characteristics, which may bound our understanding of the data by what an individual algorithm acknowledges. We believe a comprehensive toolbox like \Rdimtools helps users to grasp the nature of complex data by leveraging fragmented knowledge that multiple algorithms elaborate separately.

\section{Related Work}

Many libraries have been proposed to provide a number of algorithms in a unified framework of its own, including \texttt{drtoolbox} \citep{vandermaaten_dimensionality_2009a} in MATLAB, \texttt{scikit-learn} \citep{pedregosa_scikitlearn_2011} in Python, and a \cpp template library \texttt{tapkee} \citep{lisitsyn_tapkee_2013} with a known basis of popularity. In R, packages \texttt{dimRed} \citep{kraemer_dimred_2018}, \texttt{dyndimred} \citep{cannoodt_dyndimred_2020}, \texttt{intrinsicDimension} \citep{johnsson_structures_2016}, and others can be comparable although their scopes are not alike \Rdimtools as summarized in Table \ref{tab:comparison}.

\begin{table}[h]
	\centering
	\begin{tabular}{|c|c|c|c|c|c|}
		\hline
		\multirow{2}{*}{Library} & \multirow{2}{*}{Version} & \multirow{2}{*}{Written in} & \multicolumn{3}{c|}{Number of functions}           \\ \cline{4-6} 
		&                          &                             & DR & IDE & total \\ \hline \hline
		\texttt{Rdimtools} & 1.0.2 & R & 133 & 17 & 150 \\ \hline 
		\texttt{dimRed} & 0.2.3 & R & 19 & 0 & 19 \\ \hline
		\texttt{dyndimred} & 1.0.3 & R & 12 & 0 & 12 \\ \hline
		\texttt{intrinsicDimemsion} & 1.2.0 & R & 0 & 8 & 8 \\ \hline
		\texttt{tapkee} & 1.0.2 & \cpp & 18 & 0 & 18 \\ \hline
		\texttt{drtoolbox} & 0.8.1b & MATLAB & 34 & 6 & 40 \\ \hline 
		\texttt{scikit-learn} & 0.22.2 & Python & 31 & 0 & 31 \\ \hline		
	\end{tabular}
	\caption{Numbers of algorithms supported by libraries. For \texttt{scikit-learn}, functions from linear models and feature selection modules are not counted.}
	\label{tab:comparison}
\end{table}

\section{Dependencies and Development}
\begin{figure}[h]
	\centering
	\includegraphics[width=0.75\linewidth]{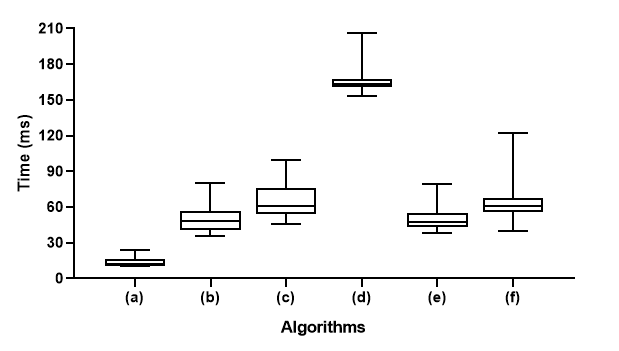}
	\caption{Runtime comparison of principal component analysis on 0.1 million samples in 72 dimensions. R packages compared are (a) \Rdimtools, (b,c) \texttt{stats} , (d) \texttt{psych}, (e) \texttt{dimRed}, and (f) \texttt{dyndimred}. Functions (b) \textsf{prcomp()} and (c) \textsf{princomp()} from \texttt{stats} package employ singular value and spectral decomposition respectively.}
	\label{fig:computation}
\end{figure}

\Rdimtools internalizes most of capabilities via a balanced mixture of R and \cpp. Below is the list of R packages upon which \Rdimtools depends.
\begin{itemize}
	\item \texttt{CVXR} \citep{fu_cvxr_2018} solves a semidefinite program and a sparse regression problem with complex constraint in 3 DR functions.
	\item \texttt{RSpectra} \citep{qiu_rspectra_2019} simplifies large-scale spectral decomposition when we only need the $k$ largest or smallest eigenpairs. 
	\item \texttt{RcppDE} \citep{eddelbuettel_rcppde_2018} performs black-box optimization via differential evolution in 2 functions to find an optimal set of parameters. 
	\item \texttt{Rcpp} \citep{eddelbuettel_rcpp_2011, eddelbuettel_seamless_2013} enables convenient integration of \cpp codes with the binding of \texttt{Armadillo} \cpp linear algebra library \citep{sanderson_armadillo_2016a} via \texttt{RcppArmadillo} \citep{eddelbuettel_rcpparmadillo_2014a}. Computational gain from \cpp is shown in Figure \ref{fig:computation}.
\end{itemize}

\Rdimtools follows modern convention of open source software development. The project is hosted on GitHub\footnote{\url{https://github.com/kyoustat/Rdimtools}} for collaborative development and each commit to the repository triggers a check to secure completeness via a continuous integration service\footnote{\url{https://travis-ci.org/github/kyoustat/Rdimtools}}. 

\Rdimtools is also available from Comprehensive R Archive Network (CRAN)\footnote{\url{https://CRAN.R-project.org/package=Rdimtools}} for easy installation and use. Distribution through CRAN mandates to include working examples for every function after checks. It plays a role of integration testing when the package is updated. Also, CRAN requires compatibility with major operating systems.\footnote{A list of CRAN check settings is at \url{https://CRAN.R-project.org/web/checks/check_flavors.html}.} 

Documentation of API can be accessed from one of the followings; (1) \textsf{help()} function from R console, (2) reference manual from CRAN, and (3) reference section from the package website\footnote{\url{https://kyoustat.com/Rdimtools}}, all of which deliver minimal working examples for families of algorithms.

\section{Functionalities of \texttt{Rdimtools}}

%

Functions in \Rdimtools are categorized into families of \textsf{do.\{algorithm\}}, \textsf{est.\{algorithm\}}, and \textsf{aux.\{algorithm\}} for DR, IDE, and auxiliary algorithms respectively. A family of DR algorithms can be further categorized as in Table \ref{tab:taxonomy}.

\begin{table}[ht]
	\centering
	\begin{tabular}{|c ||c|c|c||c|}
		\hline 
		& supervised & semi-supervised & unsupervised & total \\ \hline \hline
		linear & 43 & 5 & 46 & 94\\ \hline 
		nonlinear & 7 & 2 & 30 & 39 \\ \hline \hline 
		total & 50 & 7 & 76 & 133 \\ \hline
	\end{tabular}
	\caption{Numbers of algorithms for dimension reduction per class categorized by projection type and use of label or auxiliary information.}
	\label{tab:taxonomy}
\end{table}

\begin{figure}[h]
	\centering
	\includegraphics[width=0.9\linewidth]{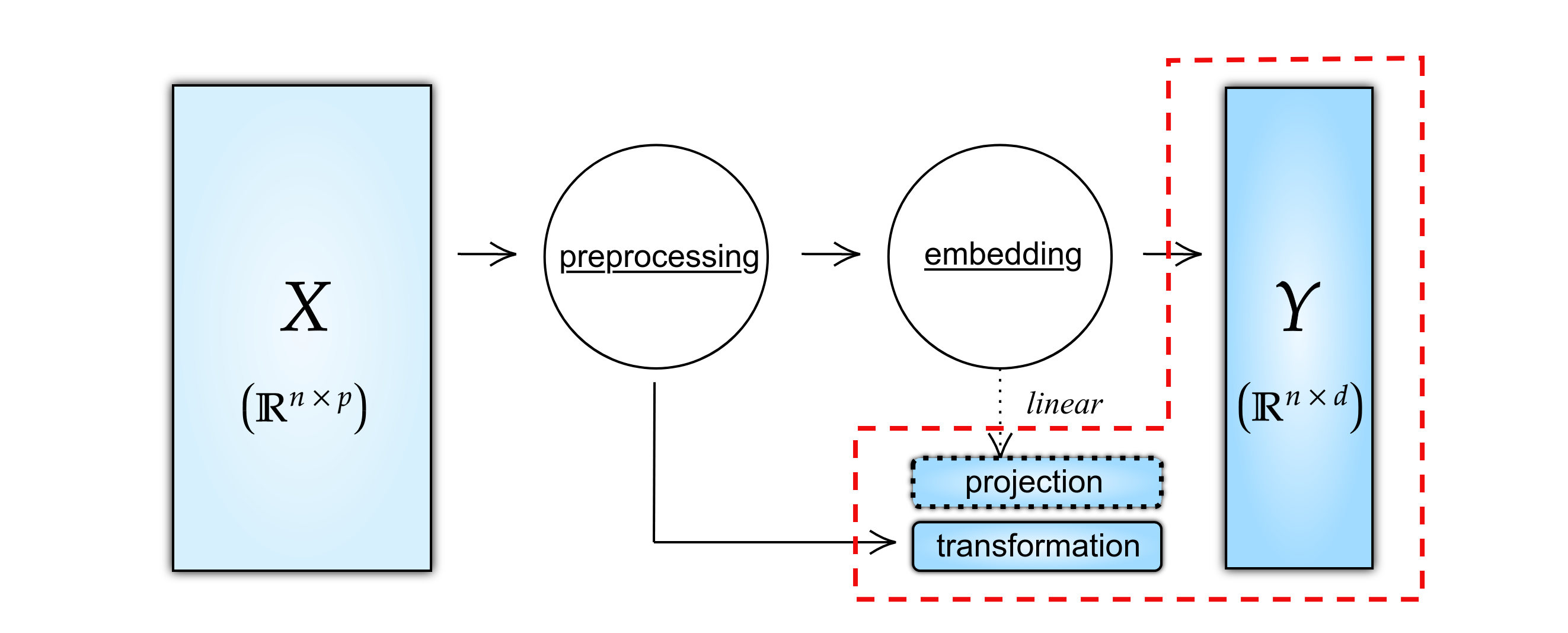}
	\caption{Common structure of DR functions. Given the data $X\in\mathbb{R}^{n\times p}$ and parameters including the target dimension $d$, a DR function returns an embedding $Y\in\mathbb{R}^{n\times d}$, preprocessing information, and projection matrix if an algorithm is of linear type.}
	\label{fig:flowchart}
\end{figure}

We describe a common structure of DR functions as shown in Figure \ref{fig:flowchart}. Given the multivariate data matrix $X \in \mathbb{R}^{n\times p}$ where rows are $p$-dimensional observations, preprocessing comes in first if applicable. The \textsf{aux.preprocess()} routine provides 5 different operations; \textsf{`center'}, \textsf{`scale'}, \textsf{`cscale'}, \textsf{`decorrelate'}, and \textsf{`whiten'}. Transformation is saved in an R list for future use. An algorithm is applied to the transformed data and returns projected coordinates $Y \in \mathbb{R}^{n\times d}$ for a predefined target dimension $d < p$. If an algorithm is of linear type, a projection matrix in $\mathbb{R}^{p\times d}$ is also returned. When an algorithm is one of 16 linear methods that employ feature selection, it also returns an index vector of $d$ variables that are selected.


17 IDE algorithms all return an estimated dimension \textsf{estdim} while methods that employ bottom-up estimation schemes also report a length-$n$ vector of local estimates at each point. 

Other notable auxiliary functions include \textsf{aux.gensamples()} to generate samples from 10 popular data models, \textsf{aux.graphnbd()} to construct $k$- and $\epsilon$-nearest-neighbor graphs that are used to approximate a data manifold embedded in $\mathbb{R}^p$, \textsf{aux.kernelcov()} to compute a positive definite kernel matrix $K(x_i,x_j) = \langle \phi(x_i),\phi(x_j) \rangle$ using 20 types of kernels \citep{hofmann_kernel_2008}, and \textsf{aux.shortestpath()} that implements Floyd-Warshall algorithm \citep{floyd_algorithm_1962a} to find shortest-path distances on a graph that approximate every pairwise geodesic distance on a data manifold reconstructed by a nearest-neighbor graph.

\section{Conclusion}

\Rdimtools puts an unprecedented number DR and IDE tools for high-dimensional data analysis in a single R package. We scratch  couple R with \cpp for fast, flexible development and efficiency. The package is maintained and distributed via CRAN, GitHub, and a package website to secure easy access, continuous integration, transparency, and collaborative development. All venues contain examples and a full API documentation.

We plan to further develop the \Rdimtools package to incorporate more algorithms and out-of-memory support in response to the needs for big data analysis. Another direction of development in progress is to translate all subroutines written in R into pure \cpp codes. This has proven to be successful in reducing communication costs of complicated algorithms. In the long run, we hope the latter effort opens up an opportunity for \Rdimtools project to evolve into a standalone \cpp library for wider use.


%
%
%
%
%
%
%
%

\vskip 0.2in
\bibliography{references}

\end{document}